\documentclass[]{ceurart}

\usepackage{csquotes}

\usepackage{xspace}
\usepackage{misc-clean}

\usepackage{tikz}
\usetikzlibrary{shapes,snakes}

\usepackage{algorithm2e}

\begin{document}

\copyrightyear{2023}
\copyrightclause{Copyright for this paper by its authors.
  Use permitted under Creative Commons License Attribution 4.0 International (CC BY 4.0).}

\conference{KBC-LM'23: Knowledge Base Construction from Pre-trained Language Models workshop at ISWC 2023}

\title{Towards Ontology Construction with Language Models}

\author[1, 2]{Maurice Funk}[%
email=maurice.funk@uni-leipzig.de%
]
\address[1]{Leipzig University}
\address[2]{Center for Scalable Data Analytics and Artificial Intelligence (ScaDS.AI)}

\author[1,3]{Simon Hosemann}[%
  email=simon.hosemann@uni-leipzig.de%
]
\address[3]{School of Embedded Composite Artificial Intelligence (SECAI)}

\author[4]{Jean Christoph Jung}[%
email = jean.jung@tu-dortmund.de%
]
\address[4]{TU Dortmund University}

\author[1, 2, 3]{Carsten Lutz}[%
  email=carsten.lutz@uni-leipzig.de%
]

\begin{abstract}
  We present a method for automatically constructing a concept
  hierarchy for a given domain by querying a large language model.  We
  apply this method to various domains using OpenAI's GPT 3.5. Our
  experiments indicate that LLMs can be of considerable help for
  constructing concept hierarchies.
\end{abstract}

\maketitle

\section{Introduction}

%
Ontologies are formal representations of the concepts in a domain and
their relations and thus represent highly structured
knowledge. However, their manual construction and curation is a
difficult engineering task that is both time consuming and costly.
This has led to the proposal of various approaches to (semi-)automatic
ontology construction, see e.g.\ the
surveys~\cite{DBLP:journals/ki/Ozaki20,DBLP:conf/semweb/VrolijkRVMMR22}.
A particular challenge is that expertise on ontology engineering and
domain knowledge are typically not in the same hands.
This has been addressed by the design of algorithms that systematically
ask questions to a domain expert and construct the ontology based on
the answers given.  Notable examples include exact learning of
ontologies in the style of Angluin
\cite{DBLP:journals/jmlr/KonevLOW17} and the use of algorithms from
formal concept analysis
\cite{DBLP:conf/ijcai/BaaderGSS07,DBLP:conf/iccs/RudolphVH07}.

While such approaches look good on paper, we are not aware that they
have been applied in practice. An obvious problem is that the domain
expert is forced into a monotonous practice of answering uninteresting
questions without knowing their exact purpose. Moreover, the expert
still needs to invest considerable time.  One may argue, however, that
with the advent of large language models (LLMs) trained on huge
corpora such as OpenAI's GPT \cite{gpt3,
  DBLP:journals/corr/abs-2112-11446,DBLP:journals/corr/abs-2204-02311,DBLP:journals/corr/abs-2201-11990},
we have available `experts' on many domains that do not easily become
tired of answering questions and that are rather affordable. In fact,
LLMs latently contain a significant body of knowledge and starting
from \cite{DBLP:conf/emnlp/PetroniRRLBWM19}, there has been a quickly
growing literature on exploiting this fact: LLMs have been used
directly as knowledge bases, for general question answering, and to
complete knowledge graphs such as Wikidata. To the best of our
knowledge, however, none of the existing studies considers ontology
construction.

The aim of this paper is to take a first step towards the
(semi-)automatic construction of ontologies based on LLMs.  Our approach
is not based on existing methodologies such as exact learning or
formal concept analysis, but specifically tailored towards LLMs. One
main reason is that existing methods assume that the schema of the
ontology (the set of concept and property names to be used) is chosen
in advance and then provided as an input to the methodology. This,
however, does not appear to be a good choice when working with LLMs,
for at least two reasons.  First, designing a schema for an entire
domain is a non-trivial task itself that requires a domain expert and
involves many design decisions. In fact, designing a schema and a
concept hierarchy are closely entangled. And second, a main strength
of LLMs is to generate keywords and phrases in a context provided by a
user and thus they are a perfect tool for proposing concept and
property names for a given domain. It seems very natural to take
advantage of this powerful feature.


The more expressive the ontology language, the more design decisions
have to be taken during ontology construction. This leads us to start,
in this initial paper, with a very simple `backbone' of ontological
representation: we only aim to construct a concept hierarchy for a
given domain, that is, we only consider the subconcept/is-a relation,
but no other relations. Our algorithm takes a seed concept $C_0$
(e.g., \textsf{Animals}), that determines the domain in the sense that
all concepts in the generated hierarchy will be subconcepts of $C_0$
(e.g., \textsf{Mammals}, \textsf{Fish}, \textsf{Lion}, \ldots). We
then `crawl' the hierarchy by repeatedly asking the LLM to provide
relevant subconcepts of concepts that are already in the hierarchy and
use an established traversal algorithm to place the new
concepts---note that each concept may have more than one superconcept
and the ultimately constructed hierarchy does not take the form of a
tree, but that of a directed acyclic graph (DAG). We also implement a
mechanism for verifying the output of the LLM by posing additional
queries to the LLM. Further, we ask the LLM to provide a textual
description of each concept that we make available for
inspection. 

To test the feasibility of our method, we apply it to various domains
such as \sf{Animals}, \sf{Drinks}, \sf{Music}, and \sf{Plants}. As the
LLM, we use GPT 3.5. A metric evaluation of the precision and recall
of the constructed ontologies is difficult because there is no ground
truth. For the time being, we thus confine ourselves to a purely
subjective evaluation based on manual inspection of the constructed
ontologies.  We believe that they are quite reasonable and demonstrate
the utility of LLMs for constructing ontologies.
%
Hallucinations and errors occur, but they can be significantly reduced
by verification and careful prompt engineering. Incompleteness also
occurs, but seems to be outweighed by the fact that our approach is
able to suggest a wealth of classes relevant to a domain as well as
their interrelationship in terms of the is-a relation.  We make our
ontologies publicly available (without any manual post-processing) and
the reader is invited to take a look.

In the form presented here, our method is fully automatic.  We do not
claim, though, that a fully automatic approach is the solution to
ontology construction in practice. Quite to the contrary, it seems
natural and useful to also include interactions with a human domain
expert to guide the construction process. We believe that our method
can easily be extended in this direction. We also believe that our
experiments indicate that involving LLMs in ontology construction can
bring about significant benefits compared to a purely manual approach,
in a similar way in which using ChatGPT can bring significant benefits for
writing text. In particular, the LLM can propose relevant concept
names with ease and also make useful suggestions regarding their
position in the hierarchy.

\paragraph{Related Work.} When the strength of LLMs increased, it
became evident that they (latently) store a massive amount of
knowledge which suggests their use as a knowledge source in
applications such as knowledge graph completion, ontology completion,
and open domain question answering.  Starting
from~\cite{DBLP:conf/emnlp/PetroniRRLBWM19}, there is a quickly
developing line of work that explores the use of LLMs for open domain
question answering, typically using `fill-in-the-blank' cloze
statements, often in the form of `subject-relation-object' triples
with a
blank~\cite{DBLP:conf/emnlp/ShinRLWS20,DBLP:conf/emnlp/RobertsRS20,DBLP:conf/naacl/QinE21,DBLP:journals/corr/abs-2208-11057,DBLP:conf/eacl/HavivBG21}.
In the same spirit but closer to our paper
is~\cite{DBLP:conf/eacl/CohenGBG23} which uses a crawling approach to
extract a knowledge graph from an LLM using the same kind of
statements. In contrast to our work, however, there is no special
focus on concept hierarchies. There seems to be only little work on
using LLMs for completing knowledge graphs or
ontologies~\cite{DBLP:conf/esws/VeseliSRW23}. Notable exceptions are
\cite{LIU2020103607} and \cite{chen2023contextual} which use
fine-tuned BERT models for subsumption prediction with the aim of
completing ontologies. This is similar to the insertion of newly
discovered concepts into the hierarchy in our approach, but it lacks
the concept discovery / crawling aspect of our
work. 

\section{The Algorithm}

A \emph{concept hierarchy} is a preordered set, that is, a pair
$(H,\sqsubseteq)$ with $H$ a set and $\sqsubseteq$ a reflexive and
transitive relation on $H$. The relation `$\sqsubseteq$' is also
called the \emph{is-a} relation or the \emph{subsumption} relation. If
$C \sqsubseteq D$, we call $C$ a \emph{subconcept} of $D$ and $D$ a
\emph{superconcept} of $C$. Note that we do not demand antisymmetry,
and thus for distinct $C,D \in H$ it is possible that
$C \sqsubseteq D \sqsubseteq C$. We then call $C$ and $D$
\emph{synonyms}.  One may equip concept hierarchies with a
set-theoretic semantics as used for example in description logics and
in OWL, but this is not necessary for the purposes of the current
paper. It often makes sense to think of the subsumption relation in
terms of its transitive reduction (also called the Hasse diagram),
which is a directed acyclic graph (DAG). 

The general strategy that we use to construct concept hierarchies from
LLMs is displayed as Algorithm~\ref{exalg}.
\begin{algorithm}[t]
  \SetAlgoLined  
  \DontPrintSemicolon  
  \SetAlgoNoEnd
  \SetKwComment{tcc}{(``}{'')}
  \SetCommentSty{} 
  \KwIn{Seed concept $C_0$}
  $H = {}$ concept hierarchy that only contains $C_0$\;  
  \While{there is an unexplored concept $C$ in $H$}{
    ask LLM whether $C$ has subconcepts\tcc*{ existence}
    \If{yes}{
      ask LLM to provide list $L$ of subconcepts of $C$\tcc*{listing}  
      ask LLM to provide descriptions of the concepts in $L$\tcc*{description}
      \ForAll{$D \in L$}{
        ask LLM to verify that $D$ is a subconcept of $C$\tcc*{verification}
        \If{successful}{
          insert $D$ into $H$\tcc*{insertion}
        }
      }
    }
  }
  \Return{$H$}
  \caption{Concept Hierarchy Construction}
  \label{exalg}
  \end{algorithm}
  The algorithm takes as input a seed concept $C_0$ that determines
  the domain for which we want to construct a concept hierarchy. For
  example, one might use here \emph{Animals}, \emph{Activities},
  \emph{Artists}, \emph{Music}, or even \emph{Things}. The algorithm
  then explores every concept $C$ that was placed in the concept
  hierarchy, starting with $C_0$, by identifying subconcepts and
  inserting them into the hierarchy. We also ask the LLM to provide
  a textual description of each concept and 
  %
  use a verification step to filter out erroneous
  answers. 
  All this is described in detail below. Note that we do not
  (additionally) traverse the concept hierarchy upwards by asking also
  for superconcepts when exploring a concept $C$ from $H$. Doing so
  bears the risk of leaving the domain, though one might invent
  measures to prevent this. The algorithm may terminate naturally
  if at some point no more concepts are proposed by the LLM, but
  there are no guarantees.

\section{Existence / Listing / Description / Verification}
\label{sec:existence}

We describe our implementation of existence, listing, description, and
verification. Insertion is discussed in
Section~\ref{sec:insertion}. To give a first impression, here are the
central phrases used in the prompts for existence, listing and description:
\begin{itemize}

\item Subconcept existence: ``\emph{Are there any generally accepted subcategories of $C$? Answer only with yes or no.}''

\item Subconcept listing: ``\emph{List all of the most important
  subcategories of $C$. Skip explanations and use a comma-separated
  format like this: important subcategory, another important
  subcategory, another important subcategory, etc.}''

\item Concept description: ``\emph{Give a brief description of
    every term on the list, considered as a subcategory of $C$, without the use of examples, in the following form:  List element 1: brief description for list element 1.
 List element 2: brief description for list element 2. $\dots$}.''

\end{itemize}
Of course, there are many natural variations of these phrases. In
particular, there are obvious alternatives for the word `subcategory'
such as subconcept, subclass, type, and so on. Changing the phrase has
an impact on the results (as almost every reformulation of a prompt)
and based on sampling various examples we decided that subcategory
gave the most convincing results.

\smallskip

To increase the completeness of the constructed hierarchies, our
approach to concept listing is actually more intricate than just using
the prompt given above. Ideally, we would like to consider all (or at
least a large number of) answers to that prompt and then include the
ones with the highest probabilities, up to a certain threshold. While
LLMs in principle provide this information, it is not accessible via
the GPT API that we use in our implementation. We therefore resort to
a frequency analysis, meaning that we pose the above prompt to GPT
many times and then take all answers that are returned with a certain
minimum frequency. As this is potentially quite costly, we implement
it in a slightly different way. We set the max$\_$tokens parameter to
1, meaning that we only ask for the \emph{first token} of an answer to
the above prompt to be returned.\footnote{Note that GPT cost depends
  on the number of tokens in the input and output.}  We then pose the
prompt many times (we choose 100) and take all tokens that are
returned with a certain minimum frequency (we choose the frequency
threshold between 5 and 20, out of 100). For each of the tokens $t$ that
surpasses the threshold, we once more ask the subconcept listing prompt
from above, extended with the sentence ``\emph{Start your answer with
  ``$t$''}$\,$''. The list of subconcepts is then taken to be the union
of the lists returned to these prompts. More information,
especially on how to set other parameters (which is crucial) is given
in Section~\ref{sect:results}.

\smallskip


Let us discuss the role of the textual descriptions that we request
from the LLM. On the one hand, we provide these as additional context
in further prompts, as described below. On the other hand, the descriptions
can also be very useful for a human user to interpret the concepts proposed
by the LLM. With the seed concept {\sf Drinks}, for example, GPT
identified (among many others) the concepts {\sf chocolate porters}
and {\sf chocolaty porters}. While these concepts may look like
synonyms, the descriptions produced by GPT reveal that chocolate
porters are porters to which some form of chocolate or cocoa has been
added during the brewing process while this is not true for chocolaty
porters which only exhibit an aroma that is reminiscent of
chocolate. A user may then decide whether this distinction is really
needed and whether both classes are relevant and should be
kept.\footnote{All our examples are ``real'' in the sense that they
  occurred during interactions with GPT. They are, however, not
  necessarily part of the concept hierarchies that we provide along
  with this paper as they might have been encountered when running
  earlier versions of our algorithm.}

When using basic forms of prompting for subconcept existence and
listing, a number of issues arise. In the following, we try to
categorize the most important types of errors:
\begin{itemize}

\item \emph{Sloppiness} / \emph{Domain Switches}. 




  The generated concept names are too abbreviated. While such a short name makes
  sense in the context of the concept for which it was returned as a
  subconcept, it does not contain enough information to stand by
  itself. When retrieving subconcepts based on short names, this often
  results in a departure from the domain set by the seed concept.

  Examples include
  $\text{\sf Tree} \sqsupseteq \text{\sf Apple} \sqsupseteq \text{\sf
    IPad}$,
  $\text{\sf Reusable Bottle} \sqsupseteq \text{\sf Glass} \sqsupseteq
  \text{\sf Tempered Glass}$, and
  ${\sf Tree} \sqsupseteq {\sf Olive} \sqsupseteq {\sf Stuffed\
    Olives}$.

  In rare cases, there are also domain switches that are unrelated to
  sloppiness. An example is
  ${\sf Drink} \sqsupseteq {\sf Water} \sqsupseteq {\sf River}$.
  

\item \emph{Attribute Inflation}. 

  Attributes are added to generate subconcepts, over and over again.
  This leads to concepts that, although not outright wrong and
  sometimes amusing, are irrelevant. Examples include {\sf Underwater
    Resource Management Games} and {\sf Customer-driven
    Scalability-focused Profit-driven Action-oriented Closing Keynote
    Speeches}.
  
\item \emph{Hallucination}. 

  The term hallucination is commonly used to refer to the tendency of
  LLMs to invent facts~\cite{hallucinations}. Here, it occurs in the
  specific form of irrelevant concepts, mostly by attribute inflation,
  as well as erroneous subconcept relations. Examples for the latter include
  $\text{\sf Non-flowering Plant} \sqsupseteq \text{\sf Fungi}$,
  $\text{\sf Moon} \sqsupseteq \text{\sf Solar Eclipse}$, and
  $\text{\sf Propositional Logic} \sqsupseteq \text{\sf Normal
    Forms}$.
  
\item \emph{Wrong Relation}. 

  Sometimes the subconcept/subcategory relation is confused with other
  relations, in particular with the `specific instance of' and `part
  of' relations. Examples for the former include
  $\text{\sf Yvy League}$
  $\text{\sf University} \sqsupseteq \text{\sf Yale University}$ and
  $\text{\sf Word Game} \sqsupseteq \text{\sf Scrabble}$. Examples
  for the latter are $\text{\sf Feet} \sqsupseteq \text{\sf Toes}$ and
  $\text{\sf Legs} \sqsupseteq \text{\sf Knees}$. This may be viewed
  as a specific form of hallucination.
  
\end{itemize}
%
In the list above, the error types are given roughly in decreasing
order of frequency with which they occurred. In fact, sloppiness and
resulting domain switches had a drastic negative effect on the quality
of the constructed hierarchies in early versions of our
algorithm. After addressing them, attribute inflation and
hallucination were the most common error types. For some domains such
as $\text{\sf Bodypart}$, `part of' occurred very often as a wrong
relation.

The central phrases for existence and listing given at the beginning
of this section have already been designed to address some error
types. In particular, the expressions ``generally accepted
subcategories'' and ``most important subcategories'' address attribute
inflation.  This, however, is not sufficient. To address errors more
properly, we use two measures: (i) further improve the prompts for
subconcept existence and listing and (ii)~concept verification. We
start with describing the former.

To address sloppiness and domain switches
when asking for existence and listing the subconcepts of some concept $C$,
we add to the prompts the seed concept $C_0$ and the superconcept $D$ of $C$
from which $C$ was first discovered.
This provides additional context and can be seen as an
instance of few-shot learning. For example, the exact prompt for
existence is:
\begin{quote}
  ``\emph{$D$ is a subcategory of $C_0$. $C$ is a subcategory of
    $D$. Are there any generally accepted subcategories of $C$? Answer
    only with yes or no.}''
\end{quote}
We have also experimented with adding information about the entire ancestry of
$C$, that
is, a complete path from $C_0$ to $C$ in the hierarchy. It seemed,
however, that this increased attribute inflation without much benefit
on sloppiness and domain switches.
To further reduce domain switches and to improve the quality of existence and
listing, we add to each prompt the textual concept description of every concept
that occurs in it.

We next describe the verification step which is intended to address
attribute inflation, instances as concepts, and domain switches.
Suppose we want to verify that $D$ is a subconcept of~$C$.
Verification consists of four steps:
\begin{enumerate}

\item Check that $D$ is not an instance.
  
    We use the prompt ``\emph{Is $D$ a specific
    instance or a subcategory of the category $C_0$? Answer only with
    Instance or Subcategory.}''

\item Check that $D$ is not a mereological part.
  
  We use the prompt ``\emph{Is $D$ a part or a subcategory of the
    category $C_0$? Answer only with Part or Subcategory.}'' 
    
\item Check that $D$ is a subcategory of the seed concept $C_0$.

  We use the prompt ``\emph{Can $D$ be considered a subcategory of
    $C_0$? Answer only with yes or no.}''

\item Check that $D$ is a subcategory of $C$.

  We use the prompt ``\emph{$C$ is a subcategory of $C_0$. Is $D$
    typically understood as a subcategory of $C$? Answer only with yes
    or no.}'' 
  
\end{enumerate}
Again, we add to each of the prompts the descriptions of all concepts
that occur in them.  The query in Point~1 turns out to be very
effective in dealing with instances as concepts and the query in
Point~2 is also effective, albeit less than the first one. If any of
the queries in Points~3 and~4 returns ``no'', it may be the case that
the concept name is too abbreviated and we make an attempt to find a
better name for the concept.  This is done using the prompt
    \begin{quote}
       ``\emph{$C$ is a subcategory of
    $C_0$. The following description outlines the characteristics of a
    subcategory of $C$. Provide a concise and unambiguous name for
    it. Provide only the name without any explanation.}''
\end{quote}
followed by the description of $D$. 
The LLM may then return a better name for the concept that passes the
verification step. Otherwise we drop $D$.



\section{Insertion}
\label{sec:insertion}

When we retrieve a new concept $C$ in the listing step, then we
already know one of its superconcepts. To properly insert $C$ into the
concept hierarchy constructed so far, however, we must know all its
super- and subconcepts among the existing concepts. We identify those
by additional queries to the LLM. In principle, this can be done in a
brute-force way by asking, for every existing concept $D$, whether
$C \sqsubseteq D$ and whether $D \sqsubseteq C$. However, this is not
practical as it easily leads to a huge number of
queries to the LLM---note that 
queries to GPT~3.5 via the OpenAI API are slow
and, when asked in large quantities, also expensive in a monetary sense.\footnote{The cost and speed
  depend on the size of the prompt and on the size of the answer. In
  our experiments, the average cost per request was \$0.0002 and each request
  took at least 0.3s, with requests that generate long answers taking several
  seconds.}

This parallels the situation of classifying a given ontology when only
a computationally expensive reasoner for deciding single subsumption
tests is available. A fundamental algorithm for this task that aims to
minimize the number of subsumption tests has been proposed in
\cite{BaaderHollunder+-KR-92}, often called the \emph{KRIS algorithm};
see also
\cite{DBLP:journals/ws/GlimmHMSS12} for improved versions. The setup in \cite{BaaderHollunder+-KR-92}
assumes that all concepts ever to be inserted into the hierarchy are
known in advance, but the algorithm also works in our case where
concept discovery and insertion alternate. We use the original KRIS
algorithm, called the enhanced traversal method in \cite{BaaderHollunder+-KR-92}, but
parallelize some subsumption tests (that is: queries to the LLM) for
improved performance.

The basic idea of the KRIS algorithm is to use, for inserting a new
concept $C$, a \emph{top search phase} to identify all superconcepts
of $C$ and a \emph{bottom search phase} to identify all subconcepts of
$C$. Both phases crucially exploit the transitivity of the subsumption
relation. The top search phase proceeds top down, meaning that it
starts at the most general concepts $D$ to check whether
$C \sqsubseteq D$ and then proceeds towards more specific $D$. The
rationale is that once a subsumption test $C \sqsubseteq D$ fails, we
do not need to test whether $C \sqsubseteq D'$ for any $D'$ with
$D' \sqsubseteq D$.  The bottom search phase is symmetric, proceeding
bottom up. There are some additional optimizations that we do not
describe here in full detail, see \cite{BaaderHollunder+-KR-92}.
The prompt that we use for testing whether $C \sqsubseteq D$ is the
same as for Query~4 in concept verification, again providing all
relevant concept descriptions.

However, 
inserting concepts this way
can introduce errors into the hierarchy. We discuss
the two most important issues.
First, querying GPT~3.5 for subcategories does not result in a
transitive subsumption relation. This is quite interesting as one
might argue that this relation, \emph{based on a language model}, is
indeed not transitive. For example, GPT~3.5 provided us with the
following relations:
$$
\begin{array}{c}
\text{\sf Commercial Building} \sqsupseteq \text{\sf Healthcare
  Facilities } \sqsupseteq \text{\sf Hospitals}
\\  
\text{\sf Commercial Building} \not\sqsupseteq \text{\sf Hospitals}.
\end{array}
$$
One explanation is that the subsumption between healthcare facilities
and commercial buildings is plain wrong. A more interesting
explanation is that GPT is US-centric and from a US perspective,
this subsumption is actually reasonable; at the same time, it is
reasonable to say that hospitals are healthcare facilities and also
that hospitals are \emph{not} (primarily) commercial buildings. The
point here seems to be that we are dealing with a language model and
language is vague and underspecified. Another example is $\text{\sf Hot
  Beverages} \sqsupseteq {\sf Coffee} \sqsupseteq {\sf Iced Coffee}$.

Obviously, accepting non-transitivity of the subsumption relation
leads the entire idea of a concept hierarchy ad absurdum: in which
sense would it still be a hierarchy? We thus deal with this issue in a
pragmatic way, essentially \emph{imposing} that the subsumption
relation is transitive. When discovering a concept $D$ as a subconcept
of a concept $C$, we take it for granted that $C \sqsubseteq E$ for
all concepts $D$ with $D \sqsubseteq E$, without verifying this using the LLM.
We also
do not depart from using the KRIS algorithm, which assumes the
subsumption relation to be transitive. If the answers given by the LLM
`are not transitive', then this may result in missing sub- and
superconcept relations.  It may, in theory, also lead to cycles in the
subsumption relation \emph{without} all concepts on
the cycle being synonyms. 
In our experiments,
however, these effects seemed to show up only rarely (for the cycles:
not at all).

\smallskip

The second important issue is related to the treatment of
synonyms. When inserting a concept~$C$, it may happen that $C$ is
classified both as a subconcept and as a superconcept of an existing
concept $D$. The KRIS algorithm then simply classifies $C$ and $D$ as
synonyms. In our algorithm, synonym detection is rather important
because the LLM may produce many small variations of the same concept
name, such as singular vs.\ plural and writing in one vs.\ two words
(``board game'' vs.\ ``boardgames''), especially when rediscovering
the same concept multiple times as a subconcept of different
superconcepts.  But it also often occurs that the answers of the LLM
wrongly identify two concepts as synonyms.
%
%
%
When we find two concepts $D_1$ and $D_2$ as candidates
to be synonyms, we thus have to analyze the situation further. We use
the following prompt:
\begin{quote}
  ``\emph{In the context of $C_0$, are $D_1$ and $D_2$ typically used
    interchangeably?  Answer only with yes or no.}''
\end{quote}
If the answer is ``yes'', we accept that $D_1$ and $D_2$ are synonyms.
If the answer is ``no'', we believe that one of $D_1 \sqsubseteq D_2$
and $D_2 \sqsubseteq D_1$ was hallucinated and ask:
\begin{quote}
  ``\emph{Consider the terms $D_1$ and $D_2$. Which of the terms is a
    subcategory of the other one? Answer in the following scheme:
    [[X]] is a subcategory of [[Y]].}''
\end{quote}
We then use the answer to resolve the situation. 
We catch a significant amount of 
hallucinated synonyms in this way, but not all. 
We also mention again at this point that the concept
descriptions provided by the LLM are very helpful for understanding
whether two concepts are synonyms or not; recall the example of {\sf
  chocolate porters} and {\sf chocolaty porters}.

\section{Results}
\label{sect:results}

We have implemented our algorithm in Python based on GPT 3.5 turbo.
We do not use the familiar chat interface to ask our queries as a
continuous conversation, but instead pose each query independently via
the chat completion endpoint (API) V1.  
Whenever possible
we parallelize calls to the API to improve performance.

We have used our approach to construct concept
hierarchies for the following seed concepts:
$$
{\sf Activities, Animals, Buildings, Diseases, Drinks, Fuels, Goats, Music, Plants}
$$
Under \url{https://www.informatik.uni-leipzig.de/kr/onto-llm/}, we provide
visual representations of the hierarchies as vector graphics for
manual inspection. 
We also provide them as OWL ontologies in the RDF/XML format for use
in ontology editors and offer a web interface for browsing. The OWL
ontologies also contain the textual descriptions of the concepts
provided by GPT~3.5.

\begin{figure}[t]
  \centering
  \begin{tikzpicture}
    \node[shape=ellipse,draw, inner sep=1pt] (g) at (0,0) {\textsf{Goats}};
    \node[shape=ellipse,draw, inner sep=1pt] (dairy) at (-7,-1.3)
    {\textsf{Dairy Goats}};
    \node[shape=ellipse,draw, inner sep=.5pt] (show) at (-4.2,-1.3)
    {\textsf{Show Goats}};
    \node[shape=ellipse,draw, inner sep=1pt] (mini) at (-1.6,-1.3)
    {\textsf{Mini.\ Goats}};
    \node[shape=ellipse,draw, inner sep=1pt] (meat) at (1.2,-1.3) {\textsf{Meat Goats}};
    \node[shape=ellipse,draw, inner sep=1pt] (fiber) at (4.1,-1.3) {\textsf{Fiber Goats}};
    \node[shape=ellipse,draw, inner sep=.5pt] (saanen) at (-7.5,-2.6) {\textsf{Saanen}};
    \node[shape=ellipse,draw, inner sep=.5pt] (toggen) at (-6,-3.2)
    {\textsf{Toggenburg}};
    \node[shape=ellipse,draw, inner sep=.5pt] (nig) at (-4.5,-2.6) {\textsf{Nigerian Dwarf}};
    \node[shape=ellipse,draw, inner sep=.5pt] (dwarf) at (-1.8,-3.2)
    {\textsf{Dwarf Nigerian}};
    \node[shape=ellipse,draw, inner sep=1pt] (mininub) at (0,-2.6)
    {\textsf{Mini.\ Nubian}};
    \node[shape=ellipse,draw, inner sep=1pt] (cashmere) at (3.6,-2.6)
    {\textsf{Cashmere}};
    \node[shape=ellipse,draw, inner sep=1pt] (boer) at (1.7,-3.2)
    {\textsf{Boer}};
    \node[shape=ellipse,draw, inner sep=1pt] (nigora) at (5.2,-3.2) {\textsf{Nigora}};

    \draw[->] (g) -- (meat);
    \draw[->] (g) -- (fiber);
    \draw[->] (g) -- (dairy);
    \draw[->] (g) -- (mini);
    \draw[->] (g) -- (show);
    \draw[->] (dairy) -- (nig);
    \draw[->] (dairy) -- (saanen);
    \draw[->] (dairy) -- (toggen);
    \draw[->] (mini) -- (dwarf);
    \draw[->] (mini) -- (nig);
    \draw[->] (fiber) -- (cashmere);
    \draw[->] (fiber) -- (nigora);
    \draw[->] (mini) -- (mininub);
    \draw[->] (meat) -- (boer);
  \end{tikzpicture}
  \caption{Excerpt from \textsf{Goats} hierarchy}
  \label{fig:goats}
\end{figure}
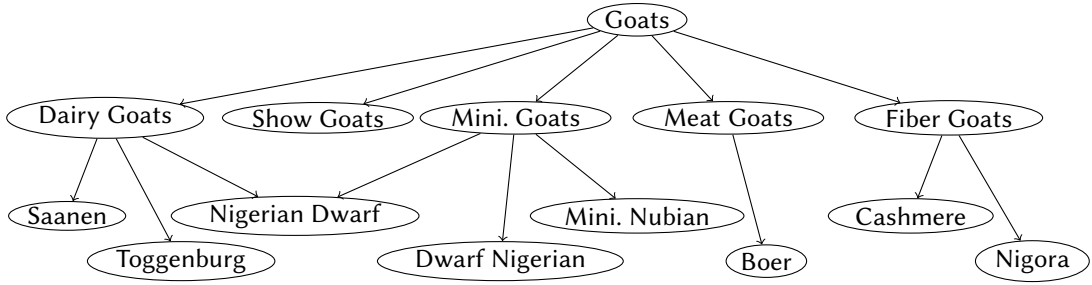

The constructed hierarchies are not perfect, but we believe that they
are quite reasonable and demonstrate the utility of LLMs for ontology
construction.
While hallucinations and errors still occur, verification and prompt
engineering have reduced them considerably.  Most of the concept names
in the hierarchy are meaningful and belong to the domain. Also the
structure of the hierarchies seems to make sense.  As a concrete
example, Figure~\ref{fig:goats} shows an excerpt of the hierarchy for
the seed concept \textsf{Goats}. It is interesting that different ways
to categorize goats play a role: by use (dairy, meat, fiber), by breed
(Nigerian Dwarf, Saanen, Boer), and by other aspects (miniature,
show). As an example for an error, note that our approach has failed
to identify \textsf{Nigerian Dwarf} and \textsf{Dwarf Nigerian} as
synonyms. The hierarchy is also incomplete. While the concept
\textsf{Miniature Nubian} was discovered and correctly placed under
\textsf{Miniature Goats}, the (arguably more important) concept
\textsf{Nubian}, a milk goat, was never discovered.

We next discuss some important parameter settings. The temperature
parameter determines the confidence that the LLM has into its most
likely predictions when choosing the next token of an answer. It 
takes values from the interval~$[0,2]$, the default being $1$. A 
value of $2$ means that the probability distribution will be very 
`flat' in the sense that many tokens get similar probabilities. At 
the other extreme, a value of $0$ means that the most likely token 
will always get chosen, resulting in almost deterministic behavior. 
The top$\_$p parameter is from the interval~$[0,1]$, the default 
being $1$, 
and it controls (on the level of tokens) which answers 
are considered at all
\cite{DBLP:journals/corr/abs-1904-09751}. If set, for example, to
$0.5$, then roughly speaking only the set of most probable tokens is
considered in which the probabilities sum up to 0.5. The interplay of
temperature and top$\_$p provides a powerful way to control GPT. In
our algorithm, we additionally have at our disposal the frequency
threshold (see Section~\ref{sec:existence}).

We generally set top$\_$p to $0.99$. Note that while this sounds
generous, it actually reduces the number of admitted tokens from
a hundred thousand down to (mostly) less than a
hundred, often less than 10. As the temperature, we choose $0$ in all
prompts except in the sampling phase of concept listing, where we set
it to~$2$. The rationale is that for sampling we want GPT to produce
as many (reasonable, whence the top$\_$p value) answers as possible
for concept listing while we want to take out randomness as much as
possible for all other prompts. Also the frequency threshold is an
interesting parameter. In our experiments, we observed a difference
between domains that have a highly structured and commonly accepted
conceptualization such as animals and domains in which
conceptualizations are less structured and more fluent, such as
activities. We refer to these as \emph{strongly structured} and
\emph{weakly structured} domains, respectively. As a rule of thumb,
lower values for the frequency threshold such as $5$ seem to work
better for strongly structured domains while higher values such as
$20$ seem more appropriate for weakly structured domains. This can be
seen as a trade-off between increased soundness (achieved by higher
values of the parameter) and increased completeness (achieved by lower
values). In the goats ontology, for instance, with the frequency
threshold of~20 that we use, no subconcepts are discovered below
\textsf{Show Goats}.  With threshold~10, subconcepts such as
\textsf{Nigerian Dwarf Show Goats} and \textsf{Toggenburg Show Goats}
appear. With threshold~5, even those have subconcepts such as
\textsf{Show Quality Nigerian Dwarf Goats} and the hallucinated
\textsf{Coat color/pattern}. We provide the Goats ontology with all
three thresholds (our favorite choice being 20) so that the reader can
get a sense for the effect of this parameter.

Regarding termination, we choose an individual exploration depth for each
domain; with the \emph{depth} of a concept $C$, we mean the length of
the shortest path from the seed concept to~$C$ in the transitive
reduction of the subsumption relation. Exploration depth $n$ means that
existence and listing is only applied to concepts of depth smaller
than $n$. The cutoff serves two purposes. On the one hand, some of the
domains have very large concept hierarchies and we want to avoid
excessive size of the provided ontologies. On the other hand, with
increasing depth (and thus increasing specificity of the concepts)
there is a tendency towards esoteric concepts. By this we mean a
concept that makes sense in principle, but has too few instances and
is too far from usual concerns to be included in the hierarchy (this
often happens via attribute inflation). This seems to occur already at
lower depths for weakly structured domains, but it happens also for
strongly structured domains when the depth increases. We could not
identify a verification/prompting strategy that stops at esoteric
concepts without also removing many non-esoteric ones. However, it is
of course easily possible to manually remove esoteric concepts after
the automatic extraction.


%
Table~\ref{tab:stat} provides statistics for the constructed
hierarchies. Column $\text{co}_d$ lists the chosen exploration depth with
`none' meaning that we run the algorithm until no more concepts were
found. ft is the frequency threshold that we have chosen, $n_C$ is the
total number of concepts, $n_D$ is the number of concepts that were
discovered but dismissed by verification, $n_{\sqsubseteq}$ is the
total number of direct subsumptions and $n_{\sqsubseteq}'$ is the
number of direct subsumptions that were discovered in the insertion
phase.  $p/C$ denotes the average number of prompts per concept and
cost denotes the overall cost of all API calls made for constructing
the hierarchy. Under $\leq \text{co}_d$ we list the number of concepts
whose depth is not larger than the exploration depth and $> \text{co}_d$ is
the number of concepts whose depth is larger. Note that concepts of
the latter kind may be introduced because the insertion phase might
place a concept below a concept that has (or exceeds) the exploration
depth. A more detailed breakdown of concept depths can be found in
Table~\ref{tab:depth} in the appendix, where we also provide
information about the outdegrees.

%
\begin{table}[t]
  \caption{Statistics for the constructed ontologies.
  }
  \label{tab:stat}
  \centering

  \begin{tabular}{cccrrrrrrrrrrr}
    \toprule
    Seed & $\text{co}_d$ &ft  & $n_C$ & $n_D$ &$n_{\sqsubseteq}$ & $n_{\sqsubseteq}'$ 
    & $p/C$ & cost (\$) & {$\leq \text{co}_d$} & { $> \text{co}_d$}
                                                          \\
                                                          \midrule 
    Activities & 3 & 20 & 545 & 227  & 969 & 425 & 38.37 & 4.07 & 273 & 272\\
    Animals & 4 & 5 & 976 & 273  & 1267 & 292 & 33.79 & 7.68 & 325 & 651\\
    Buildings & 4 & 20 & 402 & 624  & 506 & 105 & 36.30 & 2.96 & 315 & 87\\
    Diseases & 3 & 5 & 982 & 497 & 1905 & 924 & 49.89 & 11.05 & 608 & 374\\
    Drinks & 4 & 20 & 240 & 73 & 300 & 61 & 21.72 & 1.03 & 209 & 31\\
    Fuels & none & 20 & 131 & 66 & 178 & 48 & 26.01 & 0.74 & 131 & 0\\
    Goats & none & 20 & 24 & 15 & 24 & 1 & 22.25 & 0.11 & 24 & 0 \\
    Music & 2 & 20 & 453 & 89 & 735 & 283 & 46.38 & 4.30 & 266 & 187\\
    Plants & 4 & 5 & 1385 & 435 & 2473 & 1089 & 41.14 & 12.45 & 777 & 608\\
  \bottomrule
\end{tabular}

\end{table}

\section{Discussion}
\label{sect:concl}

We have presented an approach for constructing ontologies, which for now
take the form of concept hierarchies, from large language models such
as GPT~3.5. To the best of our knowledge, we are the first to do so.
We believe that there are many interesting follow-up questions to our work
that we discuss in the following.


\paragraph{Interaction with Human Domain Expert.} As discussed in the
introduction, it seems natural to add interaction with a human
domain expert to the methodology. After all, an ontology is the result
of a conscious design process. For example, reconsider
  the ontology for the seed concept {\sf Goats} depicted in
  Figure~\ref{fig:goats}.  We believe that it depends on the use case
  whether the intended direct subconcepts of {\sf Goats} are breeds
  such as {\sf Saanen} and {\sf Nigerian Dwarf} or whether they are
  related to use such as {\sf Dairy Goats} and {\sf Fiber Goats},
  potentially with the breeds as subconcepts below them. We believe
that such design decisions cannot assumed to be taken `correctly' by
the LLMs, but human intervention is required.  Another useful input
from a human user would be to control the introduction of `esoteric'
concepts.

\paragraph{Evaluation of Constructed Ontologies.}  
As there is no ground truth, already the precision of the constructed
ontologies is difficult to evaluate, and recall is even harder. In
this paper, all evaluation was purely manual and subjective. One may
think of more systematic but still manual evaluation strategies, e.g.,
via crowdsourcing. One may also try to use existing taxonomies
provided by knowledge bases such as Wikidata and Yago. An obvious
challenge is then that the concept names used by these knowledge
sources will diverge from those proposed by the LLM. It thus seems
necessary to include some form of ontology matching, which is
error-prone. In this context, it is also interesting to
  note that the hierarchies for the strongly structured domains of
  animals and plants constructed by our approach correspond more to an
  `everyday conceptualization' of these domains rather than reflecting
  scientific taxonomies. A related but different idea is to use LLMs
not for constructing ontologies, but for verifying the correctness of
existing ontologies. We are somewhat sceptical that this will bring
about good results due to the fact that most ontologies include quite
a few concepts whose names are not generally understandable. In the
medical ontology {\sc Snomed CT}, for example, there are concepts such
as ``{\sf Parameter (observable entity)}'', ``{\sf Number of pieces in
  fragmented specimen}'', and ``{\sf Counseling procedure with
  explicit context}''.


\paragraph{Philosophical Musings.} 
We believe that the constructed ontologies also raise interesting
questions from the perspective of the social sciences.  The main one
is: \emph{What do these ontologies represent?}  Since GPT~3.5 was trained
on a large fraction of human knowledge (or at least of internet
knowledge) one might ask whether the constructed ontologies represent
or approximate, at least in part, the common conceptualization of the
world shared by humanity. Note that anthropological research has found
considerable evidence that independent populations consistently arrive
at highly similar category systems across a range of basic topics, so
it is not absurd to assume that (at least to some extent) such common
conceptualizations exist, see for example~\cite{category-convergence}
and references therein. At the same time, the constructed ontologies
show a cultural bias towards the western world and, most strongly,
towards the US. For example, \textsf{Chai Tea} is a synonym of
\textsf{Spiced Tea}, which might be accepted in western countries
while in many other countries, chai is simply a synonym for tea. It
might thus also be interesting to construct ontologies in different
languages and to compare the outcome for highlighting and analyzing
the cultural impact on conceptualizations.

\paragraph{Querying GPT.} It is well-known that performance of LLMs in
knowledge acquisition tasks heavily depends on engineering good
prompts, and that small changes to the prompts can result in drastic
changes of the
output~\cite{DBLP:conf/icml/ZhaoWFK021,DBLP:conf/emnlp/HoltzmanWSCZ21}.
While we have put effort into careful prompt engineering, there is
certainly room for improvement and experimentation. Although this is
worthwhile, it is not so clear whether general lessons can be learned
from it. Will the prompts also work for other LLMs or even for the
next version of GPT? In the following, we discuss a few
aspects. 
Our prompts are mostly based on zero-shot learning, meaning that we
directly pose to the LLM the questions that we want to get answers
to. Only for existence and listing, we use a mild form of few-show
learning. It is well-known that the newest generation of LLMs is very
good at few-shot learning~\cite{gpt3}. Going beyond that,
  one could try to fine-tune the LLM towards ontology construction,
  hoping to then get by with simpler prompts. It is also an
  interesting question whether one should do prompt engineering and
  fine-tuning for \emph{domain-specific} ontology construction rather
  than in a domain-independent way.  For example, when using the seed
  concept \textsf{Animal}, we might want to ask for subspecies, rather
  than for subcategories. We mention that ontology-dependent
  fine-tuning is used in BERT-based ontology completion \cite{LIU2020103607,chen2023contextual}.

\paragraph{Querying GPT?} One may also question whether it is a good
choice in the first place to use `general-purpose' LLMs such as GPT as
an `all-domain domain expert'. To construct a high-quality ontology
for a specific domain, one might instead try to first train an LLM
specifically on selected and high-quality texts from that domain, and to
then extract an ontology from the resulting domain-specific LLM.

\paragraph{Expressive Ontologies.}
An important direction for future work is to construct ontologies that
are more expressive than concept hierarchies. There are many possible
directions. For a start, one could add disjointness constraints
between concepts. One can also extract and add instances of concepts,
which brings us closer to knowledge graph construction from LLMs, see
the related work section. Being more adventurous, one could try to
construct ontologies formulated in RDF Schema, in OWL 2 DL, or in OWL 2
QL. In all these cases, one needs to (use LLMs to) identify also
property names that are relevant for the domain under
consideration. Increasing the expressive power
brings about more modeling decisions.  For example, should a red car be
modeled as a concept {\sf RedCar}, as a conjunction
$\text{\sf Red} \sqcap \text{\sf Car}$ or even as
$\text{\sf Car} \sqcap \exists \text{\sf hasColor} . \text{\sf Red}$?
It is far from clear how such modeling decisions should be taken. 

\begin{acknowledgments}
This work is partly supported by BMBF (Federal Ministry of Education and Research) 
in DAAD project 57616814 (\href{https://secai.org/}{SECAI, School of Embedded Composite AI})
as part of the program Konrad Zuse Schools of Excellence in Artificial Intelligence.
\end{acknowledgments}

\bibliography{gpt-paper}

\appendix

\newpage
\section{Additional Statistics for Provided Ontologies}

\begin{table}[h]
  \caption{Distribution of concept depths (length of shortest path to
    seed concept).
  }
  \label{tab:depth}
  \centering

  \begin{tabular}{crrrrrrrrr}
    \toprule
    & \multicolumn{9}{c}{Number of concepts with depth $d$}\\
    \cmidrule{2-10}
    Seed & 1 & 2 & 3 & 4 & 5 & 6 & 7 & 8 & 9   \\\midrule
    Activities & 10 & 105 & \textbf{157} & 133 & 52 & 49 & 26 & 8 & 4     \\
    Animals & 2 & 16 & 57 & \textbf{249} & 324 & 219 & 83 & 21 & 4     \\
    Buildings & 13 & 73 & 121 & \textbf{107} & 54 & 11 & 11 & 8 & 3      \\
    Diseases & 23 & 228 & \textbf{356} & 267 & 95 & 12 & 0 & 0 & 0     \\
    Drinks & 4 & 24 & 60 & \textbf{120} & 25 & 6 & 0 & 0 & 0     \\
    Fuels & 7 & 29 & 34 & 32 & 25 & 3 & 0 & 0 & 0     \\
    Goats & 7 & 14 & 2 & 0 & 0 & 0 & 0 & 0 & 0     \\
    Music & 31 & \textbf{234} & 121 & 51 & 14 & 1 & 0 & 0 & 0     \\
    Plants & 8 & 65 & 220 & \textbf{482} & 380 & 151 & 57 & 17 & 4   \\
  \bottomrule
\end{tabular}

Bold values mark the exploration depth of the respective ontology.

\end{table}

\begin{table}[h]
  \caption{Distribution of outdegrees,  maximal and average outdegree.
  }
  \label{tab:outdegree}
  \centering

  \begin{tabular}{crrrrrrrrrrrrr}
    \toprule
    & \multicolumn{11}{c}{Number of concepts with outdegree $o$}\\
    \cmidrule{2-12}
    Seed & 0 & 1 & 2 & 3 & 4 & 5 & 6 & 7 & 8 & 9 & 10+ & max $o$ & avg $o$\\\midrule
    Activities & 311 & 83 & 37 & 20 & 22 & 14 & 6 & 11 & 9 & 5 & 27 & 38 & 1.78\\
    Animals & 575 & 174 & 63 & 46 & 32 & 28 & 11 & 9 & 8 & 5 & 25 & 28 & 1.30\\
    Buildings & 225 & 66 & 34 & 27 & 19 & 10 & 8 & 3 & 6 & 0 & 4 & 14 & 1.26 \\
    Diseases & 564 & 139 & 72 & 43 & 28 & 33 & 16 & 21 & 5 & 9 & 52 & 49 & 1.94\\
    Drinks & 153 & 29 & 12 & 5 & 14 & 9 & 9 & 5 & 0 & 1 & 3 & 13 & 1.25\\
    Fuels & 81 & 12 & 9 & 7 & 8 & 4 & 5 & 1 & 1 & 2 & 1 & 12 & 1.36\\
    Goats & 17 & 2 & 1 & 0 & 2 & 1 & 0 & 1 & 0 & 0 & 0 & 7 & 1.00\\
    Music & 316 & 48 & 12 & 7 & 11 & 5 & 7 & 3 & 4 & 7 & 27 & 31 & 1.62\\
    Plants & 745 & 237 & 109 & 73 & 40 & 37 & 32 & 19 & 11 & 25 & 57 & 42 & 1.79\\
  \bottomrule
\end{tabular}

\end{table}

\end{document}